\documentclass[pmlr,twocolumn,10pt]{jmlr} 
\setcitestyle{numbers}

\usepackage{rotating}
\usepackage{longtable}

\usepackage{booktabs}
\usepackage{siunitx}

\usepackage[switch]{lineno}

\usepackage[utf8]{inputenc} 
\usepackage[T1]{fontenc}    
\usepackage{hyperref}       
\usepackage{url}            
\usepackage{booktabs}       
\usepackage{amsfonts}       
\usepackage{nicefrac}       
\usepackage{microtype}      
\usepackage{xcolor}         
\usepackage{graphicx}
\usepackage{enumitem}
\usepackage{amsmath}
\usepackage{array}
\usepackage{wrapfig}
\usepackage{floatflt}
\usepackage{lipsum}
\usepackage{multirow}
\usepackage{makecell}

\hypersetup{
    colorlinks=true,
    linkcolor=black,
    citecolor=black,
    filecolor=black,
    urlcolor=black,
}

\newcommand{\ehrmamba}{{\textbf{\textsc{EhrMamba}}}}
\newcommand{\odyssey}{{\textbf{\textsc{Odyssey}}}}

\newcommand{\equal}[1]{{\hypersetup{linkcolor=black}\thanks{#1}}}

\theorembodyfont{\upshape}
\theoremheaderfont{\scshape}
\theorempostheader{:}
\theoremsep{\newline}

\jmlrvolume{259}
\jmlryear{2024}
\jmlrsubmitted{LEAVE UNSET}
\jmlrpublished{LEAVE UNSET}
\jmlrworkshop{Machine Learning for Health (ML4H) 2024} 

 \title[\ehrmamba: Towards Generalizable and Scalable Foundation Models for Electronic Health Records]{\ehrmamba: Towards Generalizable and Scalable\\Foundation Models for Electronic Health Records}

\author{%
\Name{Adibvafa Fallahpour}$^{1,2}$\equal{Equal contribution} \Email{adibvafa.fallahpour@vectorinstitute.ai}
\AND
\Name{Mahshid Alinoori}$^1$\footnotemark[1] \Email{mahshid.alinoori@vectorinstitute.ai}
\AND
\Name{Wenqian Ye}$^3$ \Email{wenqian@virginia.edu}
\AND
\Name{Xu Cao}$^4$ \Email{xucao2@illinois.edu}
\AND
\Name{Arash Afkanpour}$^1$\equal{Equal advising} \Email{arash.afkanpour@vectorinstitute.ai}
\AND
\Name{Amrit Krishnan}$^1$\footnotemark[2] \Email{amrit.krishnan@vectorinstitute.ai}
\AND
\centerline{
\addr $^1$Vector Institute, Canada \quad \addr $^2$University of Toronto, Canada \quad \addr $^3$University of Virginia, USA 
}
\\
\centerline{
\addr $^4$University of Illinois Urbana-Champaign, USA}
}
\begin{document}

\maketitle
\begin{abstract}
Transformers have significantly advanced the modeling of Electronic Health Records (EHR), yet their deployment in real-world healthcare is limited by several key challenges. Firstly, the quadratic computational cost and insufficient context length of these models hinder hospitals' ability in processing the extensive medical histories typical in EHR data. Additionally, existing models employ separate finetuning for each clinical task, complicating maintenance in healthcare environments. Moreover, these models focus exclusively on either clinical prediction or EHR forecasting, lacking proficiency in both tasks. To overcome these limitations, we introduce \ehrmamba, a robust foundation model built on the Mamba architecture. $\ehrmamba $ can process sequences up to 300\% longer than previous models due to its linear computational cost. We also introduce a novel approach to Multitask Prompted Finetuning (MPF) for EHR data, which enables $\ehrmamba $ to simultaneously learn multiple clinical tasks in a single finetuning phase, significantly enhancing deployment and cross-task generalization. Furthermore, our model leverages the HL7 FHIR data standard to simplify integration into existing hospital systems. Alongside \ehrmamba, we open-source \odyssey, a toolkit designed to support the development and deployment of EHR foundation models, with an emphasis on data standardization and interpretability. Our evaluations on the MIMIC-IV dataset demonstrate that $\ehrmamba $ advances state-of-the-art performance across 6 major clinical tasks and excels in EHR forecasting, marking a significant leap forward in the field.
\end{abstract}


\paragraph*{Data and Code Availability}
The MIMIC-IV dataset is publicly available \href{https://physionet.org/content/mimic-iv-fhir/1.0/}{\underline{here}} \cite{mimic-iv, mimic-iv-2, fhir, PhysioNet}.
All models were developed and evaluated in this study using the \hyperref[sec:odyssey]{$\odyssey$} toolkit.

\section{Introduction}
\label{sec:intro}
Personalized medicine is the pinnacle of healthcare innovation, and AI represents a promising solution \cite{ai_healthcare}. Central to this revolution are Electronic Health Records (EHR), which document the entire medical histories of patients in hospital visits \cite{shakey_foundations}. These records form detailed chronological sequences that include diagnoses, procedures, observations, medications, laboratory tests, demographics, and clinical notes of millions of patients over decades. With over 80\% of hospitals in the US and Canada adopting EHR systems, this extensive data provides an unparalleled resource for training EHR foundation models \cite{ehr_adoption, exbehrt}. These models hold the potential to personalize treatment plans, uncover disease patterns, detect the onset of rare illnesses, and enhance clinical predictions \cite{shakey_foundations, google_ehr, behrt}.
\smallskip

Transformer-based models, particularly variants of BERT \cite{bert}, have demonstrated remarkable capabilities in modeling EHR data and predicting clinical outcomes \cite{cehr_bert, behrt, exbehrt, medbert, hi-behrt, duett}. However, their translation to real-world clinical settings remains an open challenge. Existing models often prioritize research performance and may not adequately consider the practical constraints faced by hospitals for deploying such large-scale models \cite{google_ehr, shakey_foundations}. These include limited computational resources, data privacy regulations mandating on-premise deployments, and the need for flexible models that generalize to new tasks and integrate seamlessly with existing healthcare infrastructure. In this paper, we mainly focus on the following bottlenecks:

\paragraph{Computational Constraints.}
Deploying Transformer-based models in hospital settings is challenged by the length of EHR data. The quadratic scaling of computational and memory requirements becomes prohibitive when sequences span thousands of tokens to capture a patient’s entire medical history. Early diagnoses, noted at the beginning of these sequences, influence the interpretation of visits occurring decades later, requiring extensive context lengths to effectively capture these longitudinal healthcare dynamics. However, the computational resources required often far exceed what is available in many hospitals.

\paragraph{Finetuning Overhead.}
Finetuning EHR models for each downstream clinical predictive task, such as predicting patient mortality, presents several significant challenges. This process entails creating a separate copy of the base pretrained model and finetuning it for each specific task, leading to the simultaneous management and maintenance of multiple specialized models. Finetuning task-specific models in isolation demands considerable resources and hinders their ability to integrate insights across different tasks, impairing model's reliability and performance.

\medskip
To overcome these limitations, we propose \ehrmamba, a robust foundation model designed for scalable, deployable, and generalizable EHR modeling. $\ehrmamba $ is built on the Mamba architecture and introduces several key contributions:
\paragraph{Scalability.} $\ehrmamba$ 
extends the context length by 300\% compared to previous transformer-based models, enabling the processing of longer EHR sequences and the inclusion of more comprehensive information in the sequence \cite{shakey_foundations}.
\smallskip
\paragraph{Multitask Prompted Finetuning (MPF).} We train $\ehrmamba$ using a variant of MPF \cite{mpf} for EHR data, allowing simultaneous learning of multiple clinical predictive tasks within a single finetuning phase. This approach enhances cross-task generalization, supports the learning of new tasks without modifying the model architecture, and simplifies real-world deployment in hospitals.
\smallskip
\paragraph{Dual Competence.} $\ehrmamba$ is the first model to perform both EHR forecasting, predicting future data in EHR sequences, and clinical predictive modeling, predicting patient outcomes such as mortality. This dual functionality enables comprehensive disease pattern forecasting and personalized prediction timelines, facilitating tailored treatment plans based on individual patient trajectories.

\smallskip
\paragraph{\odyssey.} $\ehrmamba$ is built using \odyssey, a toolkit designed to facilitate the development and deployment of EHR foundation models. $\odyssey $ supports gathering and processing EHR sequences using the HL7 Fast Healthcare Interoperability Resources (FHIR) standard, which simplifies integration into existing hospital systems due to its widespread adoption in healthcare settings \cite{fhir}. \odyssey comprises four key modules: data processing, model implementations, evaluation tools, and interpretation methods, providing a comprehensive framework for EHR-based machine learning research. See \hyperref[sec:odyssey]{Appx.A.} for more information on \odyssey.

\medskip
We assess $\ehrmamba $ on 6 major clinical predictive tasks using the MIMIC-IV dataset \cite{mimic-iv, mimic-iv-2, PhysioNet}. Our results show that $\ehrmamba $ achieves state-of-the-art performance while operating with significant memory and computational efficiency. Additionally, we present a patient case study highlighting EHR forecasting capabilities and interpretability methods.

\begin{figure*}[t]
  \vspace{-20pt}
  \centering
  \includegraphics[width=0.8\linewidth]{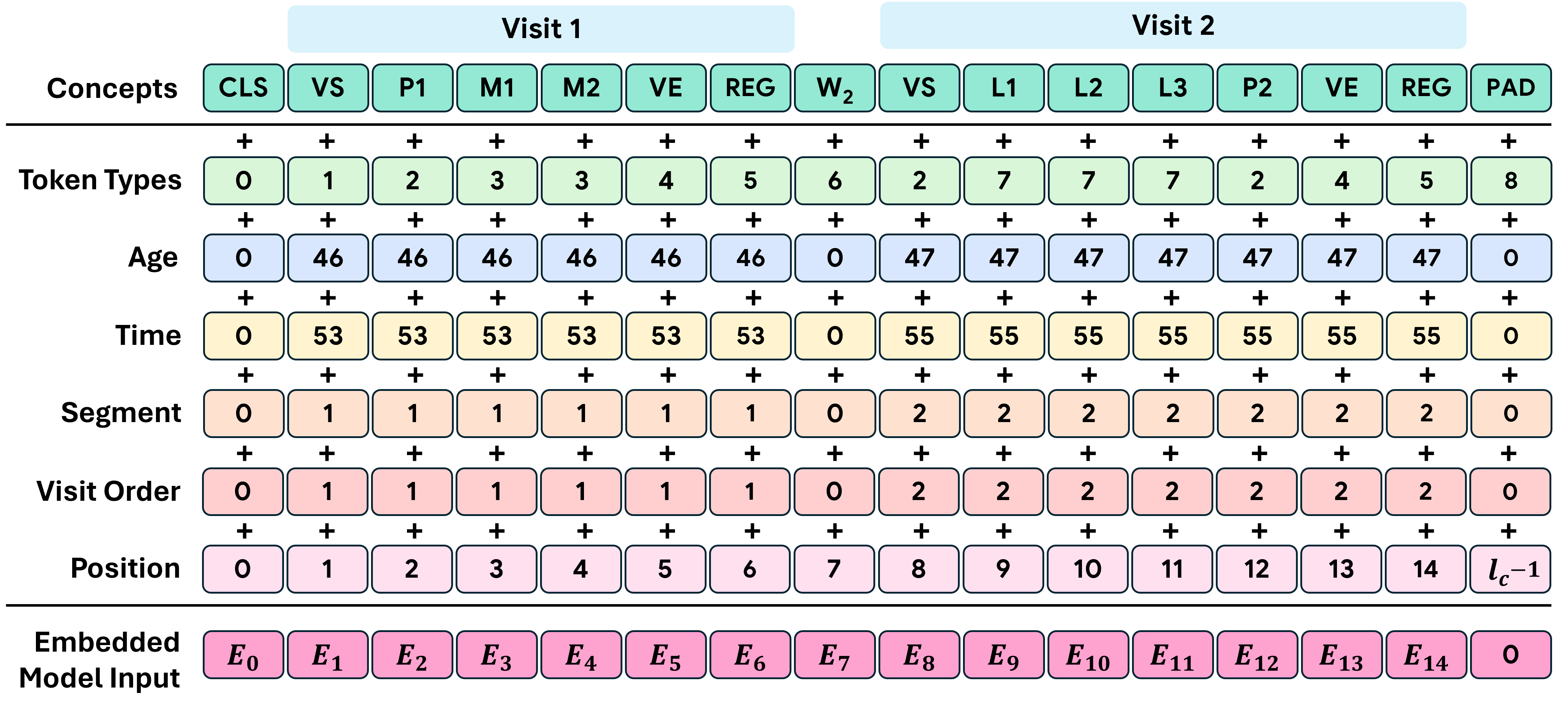}
    \vspace{-10pt}
  \caption{Patient sequence example. Visit 1 has a procedure and two medications; visit 2, after two weeks, has three lab tests and another procedure. The concept embedding of each token is added to its attribute embeddings (type, age, time, segment, visit order, position) to encode the sequence.}
  \label{fig:patient_representation}
  \vspace{-10pt}
\end{figure*}

\section{Preliminaries}
\label{Preliminaries}
\subsection{Data Representation.}
Patient sequences in EHR data represent a time series of medical events encoded using standard biomedical tokens \cite{icd, semantic_ehr, llm_ehr, medical_concept_representation}. Each patient sequence starts with a [CLS] token, similar to the classification token used in BERT \cite{bert}, followed by a series of visits marked by [VS] (visit start) and [VE] (visit end) tokens. Special tokens indicating the time interval are used between consecutive visits, e.g. [W$_2$] represents 2 weeks \cite{cehr_bert}, and each visit end token is followed by a register token [REG], similar to its use in Vision Transformers \cite{register, vit}. \hyperref[sec:appx.e.]{Appx.E.} provides more information.

Medical events within visits are organized chronologically and represented by tokens \( e_{v,j} \), where \( v \) denotes the visit number, and \( j \) is the index of the event within that visit \cite{cehr_bert}. Each token \( e_{v,j} \) is associated with a timestamp \( \tau_{v,j} \), indicating the exact time of the event such that $\tau_{v,j} \leq \tau_{v,j+1}$. We define a function \( f(e_{v,j}) : \mathcal{E} \rightarrow \Gamma \) that maps the event space \( \mathcal{E} \) to a set of attributes \( \Gamma \) including: \cite{iterative_framework}:
\begin{itemize}[left=0pt, itemsep=2pt, parsep=2pt]
    \item \textbf{\( T \)}: Type of the event, either procedure \( (P) \), medication \( (M) \), or lab result \( (L) \), see \hyperref[sec:appx.b.]{Appx.B.}
    \item \textbf{\( A \)}: Age of the patient at the time of the event.
    \item \textbf{\( \tau \)}: Exact timestamp of the event, counted in weeks relative to some reference.
    \item \textbf{\( S \)}: Visit segment, alternating between 1 and 2 to indicate the separation between adjacent visits.
    \item \textbf{\( V \)}: Visit order, the absolute visit number incrementing for each subsequent visit.
    \item \textbf{\( \mathcal{P} \)}: Absolute position of the event.
\end{itemize}
Each attribute is mapped to a distinct token space, and their embeddings are added to the embeddings of event tokens, referred to as the concept embeddings. The sequence is padded to context length \( l_c \). Figure \ref{fig:patient_representation} shows an example patient representation and the corresponding embeddings.

\smallskip
\subsection{Embeddings.}
\label{sec:embeddings}
The embedding of event token \( e_{v,j} \), denoted as \( E_{v,j} \), is the sum of its concept embedding with the respective attribute embeddings:
\begin{align}
E_{v,j} =\ & E_{\text{concept}}(e_{v,j}) + E_{\text{type}}(T_{v,j}) + E_{\text{age}}(A_{v,j}) \nonumber\\
           & + E_{\text{time}}(\tau_{v,j}) + E_{\text{segment}}(S_{v,j}) \nonumber\\
           & + E_{\text{visit order}}(V_{v,j}) + E_{\text{position}}(\mathcal{P}_{v,j})
\end{align}
Here, \( E_{v,j} \in \mathbb{R}^d \), where $d$ is embedding size \cite{behrt, cehr_bert}.

\medskip
The embeddings are learned as follows:
\begin{itemize}[left=0pt, itemsep=2pt, parsep=2pt]
    \item \textbf{Concept Embeddings}: Each token \( e_{v,j} \) maps to an embedding, like word embeddings \cite{word_embedding, word2vec}.
    \item \textbf{Token Type Embeddings}: Embeddings representing the type of medical event or special token.
    \item \textbf{Age Embeddings}: Continuous embeddings denoting age at the time of event, see \hyperref[sec:time_embeddings]{Appx.C.2} \cite{cehr_bert}.
    \item \textbf{Time Embeddings}: Continuous embeddings for elapsed time since reference, see \hyperref[sec:time_embeddings]{Appx.C.2} \cite{cehr_bert}.
    \item \textbf{Visit Segment Embeddings}: Embeddings indicating segment 1 or 2.
    \item \textbf{Visit Order Embeddings}: Embeddings representing the sequential order of visits.
    \item \textbf{Positional Embeddings}: Absolute positional embeddings \cite{bert, bigbird}.
\end{itemize}
Special tokens have their own concept, token type, and positional embeddings, while for other embeddings, a zero vector is used \cite{behrt}. This effectively captures the rich temporal and contextual information necessary for modeling patient sequences \cite{llm_ehr, semantic_ehr, time_sensitive_embedding, clinical_concept_embedding}.

\medskip
\subsection{Learning Objectives.}
The overarching goal in EHR predictive modeling is to optimize predictions of patient outcomes, such as predicting patient mortality, given patient sequences. However, given the scarcity of labeled data in healthcare settings, a two-phase learning approach is employed: First, pretraining leverages the abundant unlabeled EHR data to capture temporal dynamics and general patterns in patient sequences. Subsequently, finetuning over a small amount of labeled data adapts this prelearned knowledge to specific clinical tasks.

\medskip
\subsubsection{Pretraining.}
Pretraining can be approached in two primary ways, depending on the model type: Next Token Prediction (NTP) and Masked Language Modeling (MLM).

\paragraph{Next Token Prediction (NTP).}
In NTP, the model predicts the next event token \( e_{v,j+1} \) in a patient sequence, given all preceding tokens in the sequence, denoted as \( \mathcal{I} = \{e_{1:v,1:j}\} \):
\begin{equation}
\theta_{\text{NTP}}^* = \arg\min_\theta \mathbb{E}_{\mathcal{D}} \left[ \sum_{(v,j) \in \mathcal{I}} -\log p(e_{v,j+1} | e_{1:v,1:j}; \theta) \right]
\end{equation}
Here, \(\mathbb{E}_{\mathcal{D}}\) denotes the expectation over the empirical distribution of the dataset \(\mathcal{D}\). Optimized model parameters are denoted by \( \theta^* \).

\paragraph{Masked Language Modeling (MLM).}
In MLM, the model predicts the identity of masked event tokens, \( e_{v,j} \in \mathcal{M}\), given the unmasked tokens \( e_{\mathcal{C}(v,j)} \) as context. The set \(\mathcal{C}(v,j)\) is defined as all tokens in the sequence except the masked tokens.
\begin{equation}
\theta_{\text{MLM}}^* = \arg\min_\theta \mathbb{E}_{\mathcal{D}} \left[ \sum_{(v,j) \in \mathcal{M}} -\log p(e_{v,j} | e_{\mathcal{C}(v,j)}; \theta) \right]
\end{equation}

\smallskip
\subsubsection{\textbf{Fine-tuning.}}
In clinical predictive modeling, each task \(c_i\) from a set of \( \{c_i\}_{i=1}^K \) tasks is treated as a binary classification problem with outcomes \(y_{c_i} \in \{0, 1\}\). The traditional method finetunes model parameters \(\theta_{c_i}\) separately for each task, expressed as:
\begin{equation}
    \{\theta_{c_i}^*\}_{i=1}^K = \left\{\arg\min_{\theta_{c_i}} \mathcal{L}_i(\hat{y}_{c_{i}}(x; \theta_{c_i}), y_{c_{i}})\right\}_{i=1}^K
\end{equation}
where \(\mathcal{L}_i\) is the binary cross-entropy loss and \( \hat{y}_{c_{i}}(x; \theta_{c_i}) \) is the predicted outcome for task \(c_i\). This approach isolates the optimization for each task, preventing the model from leveraging shared knowledge across tasks. Such isolation complicates deployment and maintenance in healthcare, as each task-specific model requires separate management, increasing overhead.

\section{\ehrmamba}
\label{EHRMamba}

\subsection{State Space Models (SSMs)}
The Mamba architecture is a type of State Space Model (SSM), a powerful class of sequence models that have proven effective at handling long-range sequences \cite{mamba, s4}. SSMs map an input sequence \( x(t) \in \mathbb{R} \) to an output sequence \( y(t) \in \mathbb{R} \) via an implicit hidden state \( h(t) \in \mathbb{R}^N \), where $N$ is the state size. The mapping is defined by the following linear differential equations:
\begin{equation}
    h'(t) = Ah(t) + Bx(t), \quad\quad
    y(t) = Ch(t)
\end{equation}
Here, \( A \in \mathbb{R}^{N \times N} \), \( B \in \mathbb{R}^{N \times 1} \), and \( C \in \mathbb{R}^{1 \times N} \) are learnable matrices. For multidimensional sequences, this system is applied independently to each dimension \cite{mamba, s4, caduceus}.
\smallskip
\subsubsection{Discretization of the System}
To apply SSMs to discrete sequences, the system is discretized using a step size \( \Delta \): \cite{mamba, timeseries}
\begin{equation}
    h_t = \overline{A} h_{t-1} + \overline{B} x_t, \quad\quad
    y_t = C h_t
\end{equation}
The discrete parameters \( (\overline{A}, \overline{B}) \) are derived using the zero-order hold (ZOH) rule: \cite{mamba, s4}
\begin{equation}
    \overline{A} = \exp(\Delta A) \quad\quad
    \overline{B} = (\Delta A)^{-1} \left((\exp(\Delta A) - I\right) \cdot \Delta B \hfill
\end{equation}
\subsubsection{Execution Methods}
\paragraph{Linear Recurrence for Inference.} The model is computed step-by-step as a linear recurrence relation, as shown in Equation (6). This results in fast inference and linear scaling in relation to sequence length, enhancing deployability in resource-limited hospitals.
\paragraph{Global Convolution for Training.} The model uses a global convolution kernel \( \overline{K} \) to enable parallel sequence processing and scalable training, described by Equation (9) below: \cite{mamba, umamba, timeseries, timeseries2}
\begin{equation}
    K = (C\overline{B}, C\overline{AB}, \ldots, C\overline{A}^k\overline{B}, \ldots)
    \quad\quad
    y = x*\overline{K}
\end{equation}

\subsection{EHRMamba Architecture}
\begin{figure}
    \centering
    \hspace{-20pt}
    \includegraphics[width=0.99\linewidth]{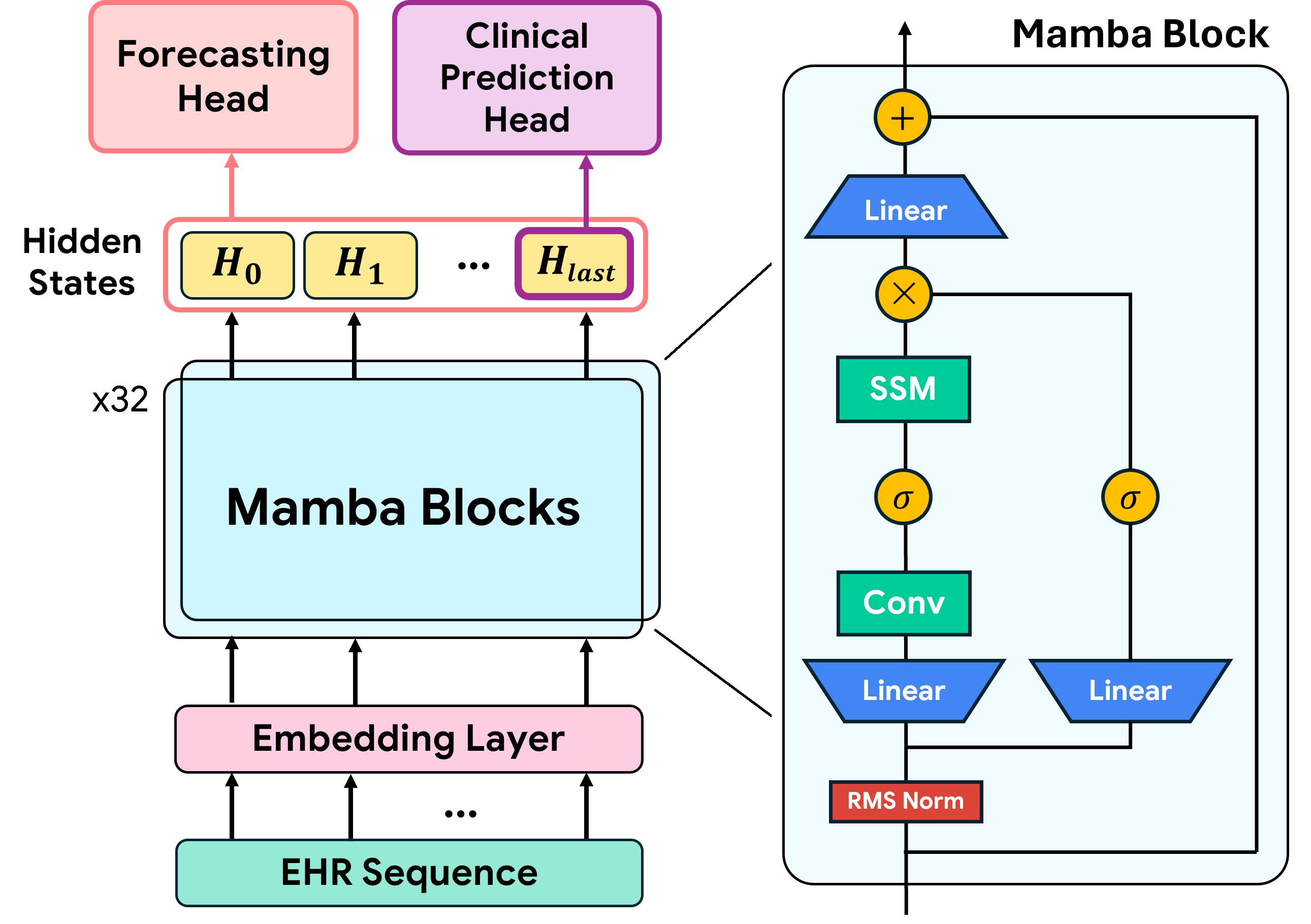}
    \caption{$\ehrmamba$ architecture. Pretraining uses the forecasting head and finetuning uses the clinical prediction head.}
    \label{fig:mamba_architecture}
    \vspace{-10pt}
\end{figure}

    EHRMamba comprises three main components: a specialized embedding layer for EHR training, multiple Mamba blocks, and two custom heads. One is used for forecasting—predicting subsequent events and visits in the patient sequence—and the other for clinical predictions, such as predicting patient mortality. Figure \ref{fig:mamba_architecture} provides an overview of the EHRMamba architecture.
    
\subsubsection{Embedding Layer}
    The embedding layer maps input sequences \( \mathbf{X} \in \mathbb{R}^{b \times l_c} \) to embedded input \( \mathbf{E} \in \mathbb{R}^{b \times l_c \times d} \) where b is the batch size. Details of the embedding layer in EHRMamba are provided in \hyperref[sec:embeddings]{Section 2.2}. Given that Mamba inherently handles sequential data, positional embeddings are excluded \cite{mamba, lstm}. 

\subsubsection{Mamba Blocks}
    The core of EHRMamba is composed of stacked Mamba blocks, each acting as a sequence-to-sequence module that preserves the input and output dimensions. These blocks map the input embeddings to the output tensor \( H \in \mathbb{R}^{b \times l_c \times d} \). Refer to \hyperref[sec:mamba_block]{Appx.C.1} for the implementation of Mamba blocks.

\medskip
\subsubsection{Forecasting Head - Pretraining}
This is the causal language modeling head that maps the last hidden states onto the vocabulary. The input first passes through a RMS normalization \cite{rmsnorm}, followed by a linear projection and softmax:
\begin{equation}
    \mathbf{\hat{Y}} = \text{\textit{Softmax}}(\text{\textit{Linear}}(\text{\textit{RMSNorm}}(\mathbf{H})))
\end{equation}
Here, $\mathbf{H}$ and \(\mathbf{\hat{Y}} \in \mathbb{R}^{b \times l_c \times v}\) represent the last hidden states and forecasting outputs, respectively, where \(v\) is the vocabulary size.

\medskip
\subsubsection{Clinical Prediction Head - Finetuning}
This is a classification head that maps $\mathbf{H}_{\text{last}}$, the last hidden state of the last token before padding in the patient sequence, to a predicted probability \( \hat{y}_c \) for a given clinical prediction task \cite{dropout, silu}:
\begin{equation}
    \mathbf{\hat{y}_c} = \text{\textit{Sigmoid}}(\text{\textit{Linear}}(\text{\textit{Dropout}}(\mathbf{H}_{\text{last}})))
\end{equation}
The Multitask Prompted Finetuning (MPF) \cite{mpf, corss_lingual_mpf} approach discussed below enables the same trained head to be used for predicting multiple clinical tasks.

\medskip
\subsection{Multitask Prompted Finetuning (MPF)}
\label{sec:mpf}
The goal of Multitask Prompted Finetuning (MPF) is to enable a single finetuned model to excel across multiple clinical tasks \cite{mpf, corss_lingual_mpf}. MPF involves a preprocessing step that replaces the first and last tokens of the sequence, a $\text{[CLS]}$ and $\text{[REG]}$ token respectively, with special task-specific tokens, such as $\text{[MOR]}$ for mortality prediction. Predictions are then derived from the last hidden state of the task-specific token. This allows the same patient sequence data to be used for various clinical tasks, each distinguished by its specific task token. During finetuning, patient sequences with task-specific tokens and appropriate labels are processed concurrently.

\medskip
MPF brings the following advantages:
\begin{itemize}[left=0pt]
    \item \textbf{Deployment.} MPF streamlines finetuning across multiple tasks and facilitates deployment on new tasks by simply introducing task-specific tokens. While using separate classification heads per task is possible, maintaining these isolated heads becomes challenging as the number of tasks grows.
    \smallskip
    \item \textbf{Learning.} MPF embeds task-specific information at the input level, enhancing the model's ability to tailor its processing and final hidden states to the specific task. This leads to efficient learning and enhances generalization across tasks, as the model is aware of the task context from the beginning.
    \smallskip
    \item \textbf{Compatibility.} MPF utilizes the identical model architectures used in popular frameworks like HuggingFace \cite{huggingface}. This compatibility facilitates the seamless transfer and reuse of trained weights, simplifying integration with existing systems and reducing deployment overhead \cite{corss_lingual_mpf}.
\end{itemize}


\section{Experiments}
\label{Experiments}

\subsection{Experimental Setup}
\paragraph{Dataset.} We evaluate $\ehrmamba $ on MIMIC-IV, a real-world, publicly available EHR dataset from Beth Israel Deaconess Medical Center \cite{mimic-iv, mimic-iv-2, PhysioNet}. It includes records from over 431,000 visits and 180,000 patients, featuring detailed temporal information on medical events such as procedures, medications, and lab results, along with demographic information such as age. Refer to \hyperref[sec:appx.b.]{Appx.B.} for more information on the dataset and data splits.

\paragraph{Clinical Predictive Tasks.} We assess models' performance on 6 primary clinical predictive binary classification tasks:
(1) Mortality Prediction, predicting whether a patient will pass away within one month after hospital discharge \cite{mortality, mortality2, mortality3}, (2) Length of Stay Prediction, estimating whether a patient's hospitalization will exceed one week based on the first 24 hours of admission \cite{los, google_ehr}, (3) Readmission Prediction, predicting the likelihood of a patient being readmitted within one month of the most recent discharge \cite{readmission, readmission2, readmission3}, along with predicting for three specific diagnostic conditions: (4) Condition 0 (Hypertension), (5) Condition 1 (Fluid Disorders), (6) Condition 2 (Lipoid Metabolism Disorders) \cite{cardiac_mortality, condition, condition2}. Refer to \hyperref[sec:appx.b.]{Appx.B.} for more information.

\paragraph{Evaluation Metrics.} We evaluate model performances using the Area Under the Receiver Operating Characteristic Curve (AUROC), the Area Under the Precision-Recall Curve (AUPRC), and the F1-Score metrics \cite{ehr_pretrain_benchmark, behrt, medbert, exbehrt}. We calculate averages and standard deviations by conducting experiments three times with randomized seeds, and use independent two-sample T-tests to assess the statistical significance of our results \cite{ehr_pretrain_benchmark}.

\begin{table*}[t]
\vspace{-15pt}
\centering
\caption{Performance comparison of $\ehrmamba $ against baseline models on 6 clinical prediction tasks using the MIMIC-IV dataset. These results demonstrate that $\ehrmamba $ achieves state-of-the-art performance while also offering memory and computational efficiency. An asterisk (*) indicates a significant improvement over the best baseline, with a p-value less than 0.05.}
\medskip
\label{table:performance-metrics}
\footnotesize
\begin{tabular}{@{}lcccccc@{}}
\toprule
Metric & XGBoost & LSTM & CEHR-BERT & BigBird & MultiBird & \textbf{\underline{\ehrmamba}} \\
\midrule
\multicolumn{7}{c}{\textbf{Mortality}} \\
\cmidrule(l){1-7}
AUROC & 0.956 (0.00) & 0.942 (0.003) & 0.967 (0.001) & 0.965 (0.002) & 0.968 (0.001) & \textbf{0.976 (0.005)*} \\
AUPRC & 0.795 (0.00) & 0.771 (0.002) & 0.857 (0.001) & 0.852 (0.004) & 0.863 (0.002) & \textbf{0.873 (0.002)*} \\
F1-Score & 0.633 (0.00) & 0.603 (0.002) & 0.751 (0.005) & 0.754 (0.004) & 0.770 (0.002) & \textbf{0.784 (0.002)*} \\
\midrule
\multicolumn{7}{c}{\textbf{Length of Stay}} \\
\cmidrule(l){1-7}
AUROC & 0.835 (0.00) & 0.854 (0.003) & 0.915 (0.003) & 0.914 (0.002) & 0.914 (0.002) & \textbf{0.919 (0.002)*} \\
AUPRC & 0.728 (0.00) & 0.774 (0.004) & 0.835 (0.005) & 0.837 (0.004) & 0.839 (0.002) & \textbf{0.849 (0.002)*} \\
F1-Score & 0.653 (0.00) & 0.685 (0.003) & 0.727 (0.002) & 0.728 (0.003) & 0.731 (0.001) & \textbf{0.738 (0.002)*} \\
\midrule
\multicolumn{7}{c}{\textbf{Readmission}} \\
\cmidrule(l){1-7}
AUROC & 0.677 (0.00) & 0.654 (0.006) & 0.672 (0.003) & 0.672 (0.002) & 0.681 (0.003) & \textbf{0.682 (0.002)} \\
AUPRC & 0.545 (0.00) & 0.536 (0.002) & 0.543 (0.003) & 0.545 (0.006) & 0.565 (0.004) & \textbf{0.567 (0.004)} \\
F1-Score & 0.533 (0.00) & 0.521 (0.004) & 0.531 (0.001) & 0.535 (0.002) & 0.544 (0.003) & \textbf{0.545 (0.003)} \\
\midrule
\multicolumn{7}{c}{\textbf{Condition 0: Hypertension}} \\
\cmidrule(l){1-7}
AUROC & 0.872 (0.00) & 0.854 (0.004) & 0.867 (0.002) & 0.874 (0.002) & 0.878 (0.001) & \textbf{0.888 (0.004)*} \\
AUPRC & 0.835 (0.00) & 0.825 (0.003) & 0.827 (0.003) & 0.840 (0.003) & 0.844 (0.003) & \textbf{0.850 (0.002)*} \\
F1-Score & 0.754 (0.00) & 0.744 (0.002) & 0.752 (0.003) & 0.761 (0.003) & 0.767 (0.003) & \textbf{0.772 (0.001)*} \\
\midrule
\multicolumn{7}{c}{\textbf{Condition 1: Fluid Disorders}} \\
\cmidrule(l){1-7}
AUROC & 0.884 (0.00) & 0.874 (0.003) & 0.879 (0.001) & 0.881 (0.002) & 0.886 (0.002) & \textbf{0.891 (0.001)*} \\
AUPRC & 0.801 (0.00) & 0.787 (0.002) & 0.793 (0.001) & 0.798 (0.002) & 0.805 (0.003) & \textbf{0.815 (0.001)*} \\
F1-Score & 0.717 (0.00) & 0.701 (0.002) & 0.719 (0.003) & 0.723 (0.003) & 0.728 (0.001) & \textbf{0.733 (0.002)*} \\
\midrule
\multicolumn{7}{c}{\textbf{Condition 2: Lipoid Metabolism Disorders}} \\
\cmidrule(l){1-7}
AUROC & 0.91 (0.00) & 0.893 (0.002) & 0.903 (0.000) & 0.906 (0.001) & 0.919 (0.002) & \textbf{0.928 (0.002*} \\
AUPRC & 0.785 (0.00) & 0.739 (0.002) & 0.745 (0.003) & 0.767 (0.001) & 0.788 (0.003) & \textbf{0.804 (0.003)*} \\
F1-Score & 0.724 (0.00) & 0.683 (0.001) & 0.690 (0.002) & 0.703 (0.003) & 0.731 (0.002) & \textbf{0.746 (0.002)*} \\
\bottomrule
\end{tabular}
\end{table*}

\begin{figure*}[t]
  \vspace{-15pt}
  \centering
  \includegraphics[width=0.9\linewidth]{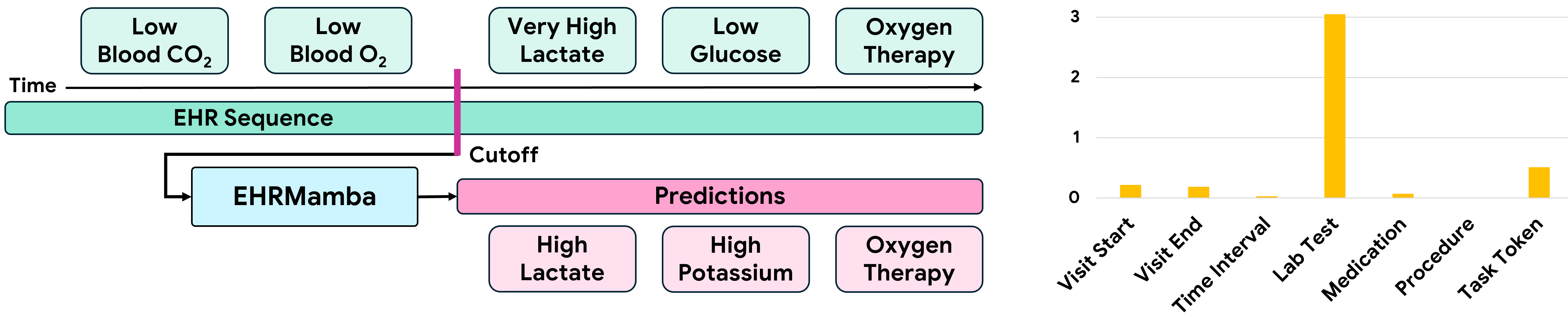}
  \caption{Left) $\ehrmamba $ predicted tokens. Right) Average token attributions for mortality prediction. Lab tests are the most influential features in $\ehrmamba$'s assessment of patient mortality risk.}
  \label{figure 3}
  \vspace{-10pt}
\end{figure*}

\smallskip
\subsection{Baseline Models.} We compare $\ehrmamba $ to 5 baseline models described below. All these models, except XGBoost, use the embedding layer describe in \hyperref[sec:embeddings]{Section 2.2}. Additionally, except MultiBird and EHRMamba, other models are trained or finetuned separately for each clinical task. Implementation and training details are provided in \hyperref[sec:appx.c.]{Appx.C.}

\paragraph{XGBoost.} The input features of the XGBoost model are frequencies of tokens from the vocabulary, excluding any special tokens, along with the age of patients in their first and last visits \cite{xgboost}. 

\paragraph{LSTM.} This is a standard BiLSTM model \cite{lstm}.

\paragraph{CEHR-BERT.} An adaptation of the BERT architecture for EHR that introduced the idea of incorporating temporal information using time embeddings and special time interval tokens \cite{bert, cehr_bert}. CEHR-BERT outperformed prior clinical BERT adaptations in various tasks including patient mortality, hospital readmission, and several disease diagnoses \cite{cehr_bert}.

\paragraph{BigBird Transformer.} The BigBird Transformer, a variant of the BERT model, employs a form of attention known as block sparse attention \cite{bert, bigbird, attention, CodonTransformer}. This modification allows for more efficient memory usage, facilitating the processing of longer EHR sequences. Here, we use a vanilla BigBird model with a context length of 2048 tokens, 4x greater than that of the CEHR-BERT model \cite{attention, cehr_bert}.

\paragraph{MultiBird Transformer.} The MultiBird Transformer adopts the structural framework of the BigBird model but is trained using MPF, see \hyperref[sec:mpf]{Section 3.3}, which enables a single finetuned model to excel across multiple downstream tasks. This training strategy is compatible with existing BigBird model implementations on Hugging Face, simplifying deployment of trained models \cite{huggingface, bigbird}.

\smallskip
\subsection{Main Results}
Table \ref{table:performance-metrics} presents the performance comparison of baseline models and $\ehrmamba$ on the MIMIC-IV dataset \cite{mimic-iv, mimic-iv-2}. EHRMamba and MultiBird show a significant performance advantage over other models, likely due to their finetuning with MPF, which enhances their ability to integrate insights across multiple tasks. This is especially beneficial for complex tasks such as readmission prediction and tasks with fewer data points, such as Condition 2. There is no substantial performance difference between BigBird and CEHR-BERT, underscoring the efficacy of block sparse attention over global attention. However, BigBird does outperform CEHR-BERT in condition prediction tasks, due to its longer context length. Notably, XGBoost struggles with tasks that require capturing temporal and sequential information, as it primarily processes token frequencies. Overall, EHRMamba outperforms MultiBird while also being far more memory and computationally efficient, making it a superior choice for a wide range of EHR modeling objectives. For a detailed comparison of the memory and computational efficiency see \hyperref[sec:efficiency]{Appx.C.3}.

\subsection{Visualized Case Study}
We present a case study of deceased Patient X, with the EHR sequence illustrated in Figure \ref{figure 3}.
\paragraph{Interpretability.} 
We use integrated gradients to assess the impact of each token in the EHR sequence on the clinical predictions. This method integrates the gradients of the model's outputs with respect to the embeddings of each input token, from a baseline (such as all zeros) to the actual input \cite{integrated_grad}. Figure \ref{figure 3} shows the average attribution scores for $\ehrmamba $'s predictions on the mortality task. Our analysis reveals that lab tests have the highest attribution values among all token types, indicating they are the most influential features for the model's mortality predictions. This implies that the model heavily relies on laboratory test results to assess patient risk.
\paragraph{Forecasting.} 
Using \ehrmamba, we forecast the subsequent event tokens in the patient X's EHR sequence given the preceding tokens before a cutoff. Figure \ref{figure 3} contrasts these predictions with actual events. Though predictions may not match exactly, they frequently capture relevant medical concepts. Appendix \ref{sec:appx.d.} offers additional examples and a quantitative evaluation of \ehrmamba's forecasting performance.

\section{Discussion}
\label{Discussion}

\paragraph{Implications and Contributions.}
$\ehrmamba$ is a significant advancement in EHR modeling. By addressing the key limitations of existing Transformer models, such as computational inefficiency and the need for separate task-specific finetuning, $\ehrmamba $ offers a scalable and deployable solution, well-suited for real-world clinical environments. The use of Mamba architecture enables processing longer patient sequences with reduced computational costs, a feasible solution for hospitals with limited resources for advanced predictive modeling. Additionally, the introduction of MPF allows $\ehrmamba $ to learn multiple clinical tasks concurrently, improving representation learning, generalizability and eliminating the overhead of maintaining multiple models.

\paragraph{Limitations.} This study is limited to the MIMIC-IV dataset with EHR data from a single hospital \cite{mimic-iv, mimic-iv-2}. Only three FHIR resources—procedures, medications, and lab tests—are used \cite{fhir}. Moreover, this dataset under represents \ehrmamba's ability to handle long EHR sequences. Our model currently does not support multimodal inputs.

\paragraph{Future Directions.} We aim to train and assess $\ehrmamba $ on a larger dataset of patient EHR records from several hospitals in Canada. We plan to include multimodal data inputs such as clinical notes and images, while expanding the context length.

\section{Related Works}
Predictive modeling using EHR data has been an active area of research, with the goal of driving personalized medicine and improving healthcare quality. Early work by Rajkomar et al. \cite{google_ehr} demonstrated the scalability and accuracy of deep learning methods on raw EHR data represented in the FHIR format, outperforming traditional predictive models for tasks like mortality prediction and diagnosis code assignment.

More recent efforts have focused on developing specialized architectures to effectively model the unique characteristics of EHR data. BEHRT \cite{behrt} was a Transformer model tailored for EHR sequences by incorporating temporal information from diagnoses, medications, and measurements. Similarly, CEHR-BERT \cite{cehr_bert} extended the BERT architecture to incorporate structured temporal EHR data in the OMOP format, improving performance on various prediction tasks. Med-BERT \cite{medbert} further adapted the BERT framework for structured EHRs, demonstrating significant improvements in disease prediction.

DuETT \cite{duett} introduced a dual event-time Transformer that models both event sequences and their corresponding timings. This temporal modeling approach showed benefits over traditional Transformers on mortality prediction and phenotyping. TransformEHR \cite{transform_ehr} proposed a denoising sequence-to-sequence encoder-decoder Transformer, pre-trained to predict all future disease diagnoses and outcomes from patient sequences. Foresight \cite{foresight} used a Generative Pre-trained Transformer for forecasting patient timelines with a variety of medical events, including diagnoses, substance use, and procedures.

\section{Conclusion}
\label{Conclusion}
We introduced \ehrmamba, a novel EHR foundation model that leverages the Mamba architecture and MPF to overcome the limitations of current Transformer models. $\ehrmamba $ excels in handling long temporal sequences and learning multiple tasks simultaneously, achieving state-of-the-art performance on six clinical prediction tasks in the MIMIC-IV dataset. Additionally, we open-sourced the $\odyssey $ toolkit which supports the development and deployment of EHR models. $\ehrmamba $ significantly advances EHR modeling, offering a robust, scalable, and generalizable solution for improving patient outcomes and clinical decision-making.


{
\clearpage
\small
\bibliography{references}
}

\clearpage
\appendix

\begin{center}
    \LARGE\bfseries Appendix
\end{center}

\section{\texorpdfstring{$\odyssey$ Toolkit}{Odyssey Toolkit}}
\label{sec:odyssey}
The $\odyssey $ toolkit is designed to support the development and deployment of EHR foundation models. It is publicly available at:
\begin{center}
    \url{https://github.com/VectorInstitute/odyssey}
\end{center}

The toolkit includes 4 major modules:
\begin{itemize}[left=0pt]
    \item \textbf{data.} This module includes scripts designed for gathering EHR datasets from HL7 FHIR resources \cite{fhir, PhysioNet}. It handles the generation and processing of patient sequences for each clinical task, tokenizing the data, and creating the necessary data splits for model training. Additionally, it provides the dataset class used for training the models.
    \medskip
    \item \textbf{models.} This module offers implementations for various models used in this study, including XGBoost, LSTM, CEHR-BERT, BigBird, MultiBird, and EHRMamba. It also includes various embedding classes essential for the models.
    \medskip
    \item \textbf{evals.} This module includes tools for testing models on different clinical prediction tasks and forecasting. It provides evaluation metrics for both training and testing data, ensuring a thorough assessment of model performance.
    \medskip
    \item \textbf{interp.} This module contains methods for interpreting model decisions. It includes interactive visualization of the attention matrix for Transformer-based models, novel interpretability techniques for EHRMamba, and gradient attribution methods. These tools enhance the transparency and understanding of model decisions.
\end{itemize}

\smallskip
\section{Data Information}
\label{sec:appx.b.}

\subsection{Dataset}
MIMIC-IV is a comprehensive EHR dataset from the Beth Israel Deaconess Medical Center, covering patient data from 2008 to 2019, including records from over 431,000 visits and 180,000 patients \cite{mimic-iv, mimic-iv-2}. The dataset is deidentified in accordance with HIPAA Safe Harbor standards and is accessible to credentialed researchers. MIMIC-IV is organized into several key tables that capture a wide range of clinical data. The most relevant components for our study include:
\begin{itemize}[left=0pt]
    \item \textbf{Procedures:} This table lists the medical procedures performed during hospital stays, such as surgeries, diagnostic tests, and therapeutic interventions. For instance, it might include details of a heart surgery or a diagnostic colonoscopy, along with when each procedure was performed.
    \medskip
    \item \textbf{Medications:} This table records medications given to patients, including drug names, dosages, methods of administration (e.g., oral, intravenous), and administration times. For example, it might show that a patient received 500 mg of Amoxicillin orally twice a day.
    \medskip
    \item \textbf{Lab Tests:} This table includes results from laboratory tests, such as blood tests, imaging studies, and other diagnostics. For example, it might list blood glucose levels or MRI results. Continuous lab test results are categorized into five bins, each represented by a unique token.
\end{itemize}

\smallskip
\subsection{Clinical Predictive Tasks}
\smallskip
\begin{itemize}[left=0pt]
\item \textbf{Mortality Prediction.} The label is determined by comparing the timestamp of the final recorded event in the patient's last visit to the timestamp of death. If the interval is less than 32 days, the outcome is labeled as 1; otherwise, it is labeled as 0 \cite{mortality, mortality2, mortality3}.
\smallskip
\item \textbf{Length of Stay Prediction.} Sequences are truncated to exclude tokens with timestamps beyond the initial 24 hours of the last visit. The sequence is labeled as 1 if the discharge timestamp exceeds one week; otherwise, it is labeled as 0 \cite{los, google_ehr}.
\smallskip
\item \textbf{Readmission Prediction.} The last visit data is excluded entirely. If the time interval between the last visit and its preceding visit is less than one month, as indicated by the special tokens for time intervals, the sequence is labeled as 1; otherwise, it is labeled as 0 \cite{readmission, readmission2, readmission3}.
\smallskip
\item \textbf{Condition Predictions.} For predicting Condition 0 (Hypertension), Condition 1 (Disorders of fluid, electrolyte, acid-base balance), and Condition 2 (Disorders of lipoid metabolism), the sequence remains unchanged. The label is assigned 1 if the condition is present, 0 otherwise \cite{condition, condition2, heart_failure_prediction}. 
\end{itemize}

\smallskip
\subsection{Data Split}
The MIMIC-IV dataset is split into pretraining (57\%), finetuning (28\%), and test (15\%) sets. The pretraining set is only utilized for MLM or NTP training, without incorporating clinical task labels \cite{ehr_pretrain_benchmark}. The finetuning set is used for EHR predictive modeling and has a greater positive ratio of task labels than the original dataset distribution. This set has no overlap with the pretraining set to ensure unbiased few shot learning evaluation. Model performance on clinical tasks and forecasting is assessed using the test set. Detailed dataset statistics are presented in Tables 2 and 3.

\smallskip
\subsection{Code and Data Access}
The MIMIC-IV dataset is publicly available at \cite{mimic-iv, mimic-iv-2, fhir, PhysioNet}:
\begin{center}
    \url{https://physionet.org/content/mimic-iv-fhir/1.0/}.
\end{center}
All models were developed and evaluated in this study using the \hyperref[sec:odyssey]{$\odyssey $} toolkit.

\section{Model Implementation and Training} 
\label{sec:appx.c.}
Here, we describe the implementation and training of models used in our experiments. All models, except XGBoost and LSTM, were trained using the AdamW optimizer \cite{adam, adamw} with a linear learning rate scheduler. During the initial 10\% of training, the learning rate increased from $5 \times 10^{-7}$ to $5 \times 10^{-5}$, and during the remaining 90\%, it decreased back to $5 \times 10^{-5}$. While XGBoost was trained on a single NVIDIA T4 16GB GPU, others utilized NVIDIA A100 80GB GPUs.

\subsection{Mamba Block}
\label{sec:mamba_block}
The input tensor \( H_{in} \in \mathbb{R}^{b \times l_c \times d} \) to each Mamba block is first normalized using root mean square layer normalization (RMS normalization) \cite{layernorm, rmsnorm}. After normalization, the tensor is expanded through two linear projections. One projection undergoes a convolution followed by a SiLU activation, denoted by $\sigma$, and its output is processed by the discretized SSM to filter relevant information. The other projection is directly passed through a SiLU activation and combined with the outputs from the SSM using a multiplicative gate. This combined output is then passed through another linear projection and summed with the initial input \( H_{in} \), resulting in the final output tensor \( H_{out} \in \mathbb{R}^{b \times l_c \times d} \). \cite{mamba, timeseries, timeseries2}

\subsection{Time Embeddings}
\label{sec:time_embeddings}
Time embeddings are designed to encode temporal information, which cannot be directly represented by standard procedures. Inspired by Time2Vec approach \cite{time2vec}, a Fourier transform is used to decompose a sequence of time points into sine functions. These functions are controlled by learnable parameters that adapt to training data. The embeddings for concept, time, and age are concatenated and projected to the original dimension, generating temporal concept embeddings \cite{cehr_bert}.

\subsection{Compute and Memory Efficiency}
\label{sec:efficiency}
EHRMamba demonstrates superior memory and computational efficiency compared to MultiBird, a transformer-based model. During pretraining, EHRMamba processed each epoch in 14:29 minutes with a maximum batch size of 44, while MultiBird required 24:05 minutes with a maximum batch size of 32. The efficiency gap widened during fine-tuning, where EHRMamba completed each epoch in 1 hour 17 minutes with a maximum batch size of 64, compared to MultiBird's 4 hours 37 minutes with a maximum batch size of 26. These results highlight \ehrmamba's ability to handle larger batch sizes and process data more rapidly, showcasing its enhanced memory utilization and computational efficiency.

\section{Forecasting}
\label{sec:appx.d.}
To evaluate \ehrmamba's predictive performance, we conducted an experiment using both train and test datasets. For each patient sequence, we set a cutoff of 10 tokens before the end and tasked the model with predicting these final 10 tokens. We then compared the predictions to the actual tokens using two metrics: accuracy and cosine similarity of token embeddings. This comparison was performed for the first 1, 2, 5, and 10 predicted tokens. The results are summarized in Table \ref{table:forcasting}.

\begin{figure*}[h]
  \begin{center}
    \includegraphics[width=1\linewidth]{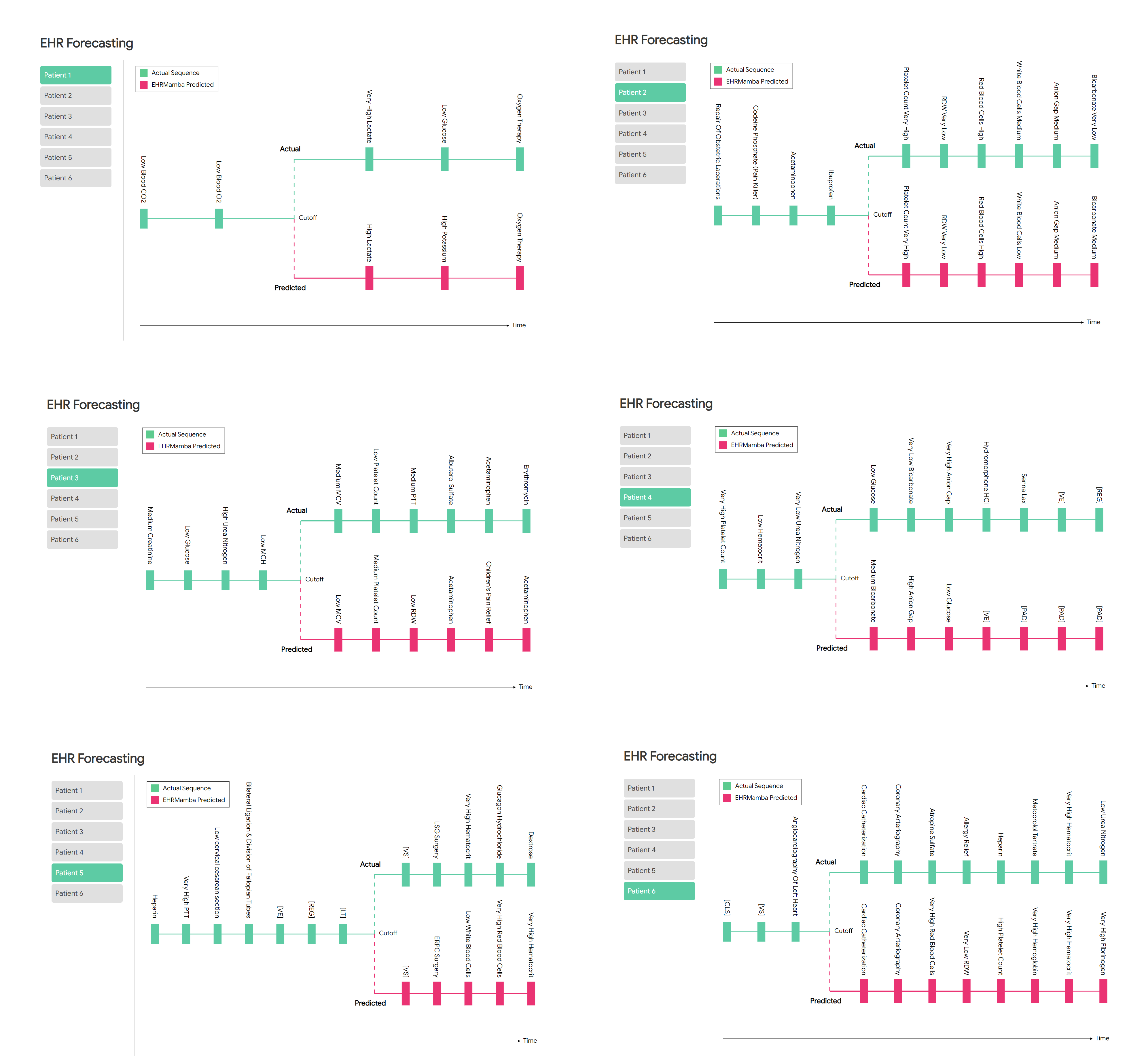}
  \end{center}
  \caption{Forecasting results on six patients of the test dataset.}
  \label{fig:forecast}
\end{figure*}

Additionally, Figure \ref{fig:forecast} illustrates the comparison between \ehrmamba's predicted EHR trajectories and the actual trajectories for six sample patients from the test dataset.

\begin{table*}[h]
\centering
\caption{Dataset Statistics for MIMIC-IV.}
\medskip
\footnotesize
\begin{tabular}{@{}lccc@{}}
\toprule
 & \textbf{Pretrain} & \textbf{Finetune} & \textbf{Test\quad} \\
\midrule
\textbf{\# Patients} & 93,787 & 47,447 & 24,924 \\
\textbf{\# Visits} & 175,413 & 137,781 & 55,589 \\
\textbf{\# Tokens} & 22,559,083 & 25,936,531 & 8,607,890 \\
\bottomrule
\end{tabular}
\end{table*}

\begin{table*}[t]
\centering
\caption{Dataset Statistics for Clinical Predictive Tasks.}
\medskip
\footnotesize
\begin{tabular}{@{}lcccccccc@{}}
\toprule
 & \multicolumn{2}{c}{\textbf{Mortality}} & \multicolumn{2}{c}{\textbf{Length of Stay}} & \multicolumn{2}{c}{\textbf{Readmission}} \\
\cmidrule(l){2-3} \cmidrule(l){4-5} \cmidrule(l){6-7}
 & \textbf{Finetune} & \textbf{Test} & \textbf{Finetune} & \textbf{Test} & \textbf{Finetune} & \textbf{Test} \\
\midrule
\textbf{\# Patients} & 47,447 & 24,924 & 42,417 & 21,595 & 31,775 & 11,338 \\
\textbf{\# Visits} & 137,781 & 55,589 & 124,889 & 49186 & 122,109 & 42,003 \\
\textbf{\# Tokens} & 25,936,531 & 8,607,890 & 24,426,766 & 8,080,243 & 20,851,581 & 6,289,346 \\
\textbf{Task Rate} & 27\% & 10\% & 47\% & 30\% & 49\% & 38\%\\
\midrule
\midrule
 & \multicolumn{2}{c}{\textbf{Hypertension}} & \multicolumn{2}{c}{\textbf{Fluid Disorders}} & \multicolumn{2}{c}{\textbf{Lipoid Metabolism Disorders}} \\
\cmidrule(l){2-3} \cmidrule(l){4-5} \cmidrule(l){6-7}
 & \textbf{Finetune} & \textbf{Test} & \textbf{Finetune} & \textbf{Test} & \textbf{Finetune} & \textbf{Test} \\
\midrule
\textbf{\# Patients} & 47,447 & 24,924 & 47,447 & 24,924 & 47,447 & 24,924 \\
\textbf{\# Visits} & 137,781 & 55,589 & 137,781 & 55,589 & 137,781 & 55,589 \\
\textbf{\# Tokens} & 20,851,581 & 8,607,890 & 20,851,581 & 8,607,890 & 20,851,581 & 8,607,890 \\
\textbf{Task Rate} & 50\% & 43\% & 49\% & 33\% & 30\% & 24\% \\ 
\bottomrule
\end{tabular}
\end{table*}

\begin{table*}[t]
\centering
\caption{Architecture and Training Details.}
\medskip
\footnotesize
    \begin{tabular}{@{}llccccc@{}}
    \toprule
     & & \textbf{LSTM} & \textbf{CEHR-BERT} & \textbf{BigBird} & \textbf{MultiBird} & \textbf{EHRMamba} \\
    \midrule
    \multirow{11}{*}{\rotatebox[origin=c]{90}{\textbf{Architecture}}} & \textbf{Embedding Size} & 768 & 768 & 768 & 768 & 768 \\
     & \textbf{Time Embedding Size} & 32 & 32 & 32 & 32 & 32 \\
     & \textbf{Visit Order Size} & 3 & 3 & 3 & 3 & 3 \\
     & \textbf{Type Vocab Size} & 9 & 9 & 9 & 9 & 9 \\
     & \textbf{Context Length} & 512 & 512 & 2048 & 2048 & 2048 \\
     & \textbf{\# Hidden Layers} & 6 & 6 & 6 & 6 & - \\
     & \textbf{\# Mamba Blocks} & - & - & - & - & 32 \\
     & \textbf{\# Heads} & - & 8 & 12 & 12 & - \\
     & \textbf{Intermediate Size} & - & 3072 & 3072 & 3072 & - \\
     & \textbf{State Size} & - & - & - & - & 16 \\
     & \textbf{Conv Kernel Size} & - & - & - & - & 4 \\
     & \textbf{Hidden Size} & 384 & - & - & - & - \\
    \midrule
    \multirow{4}{*}{\rotatebox[origin=c]{90}{\textbf{Pretrain}}} & \textbf{Batch Size} & 32 & 32 & 32 & 32 & 44 \\
     & \textbf{GPUs} & - & 4 & 4 & 4 & 4 \\
     & \textbf{Epochs} & - & 15 & 15 & 15 & 15 \\
     & \textbf{Mask Probability} & - & 0.15 & 0.15 & 0.15 & - \\
      & \textbf{Dropout Probability} & 0.1 & 0.1 & 0.1 & 0.1 & 0.1 \\
    \midrule
    \multirow{3}{*}{\rotatebox[origin=c]{90}{\textbf{Finetune}}} & \textbf{Batch Size} & 32 & 32 & 26 & 26 & 64 \\
     & \textbf{GPUs} & 1 & 4 & 4 & 4 & 4 \\
     & \textbf{Epochs} & 5 & 5 & 3 & 6 & 6 \\
     & \textbf{Dropout Probability} & 0.1 & 0.1 & 0.1 & 0.1 & 0.1 \\
    \midrule
     & \textbf{Learning Rate} & $1 \times 10^{-3}$ & $5 \times 10^{-5}$ & $5 \times 10^{-5}$ & $5 \times 10^{-5}$ & $5 \times 10^{-5}$ \\
    \bottomrule
    \end{tabular}
\label{tab:models}
\end{table*}

\begin{table*}[t]
\centering
\caption{$\ehrmamba $ Forecasting Performance}
\medskip
\footnotesize
\begin{tabular}{@{}lcccc@{}}
\toprule
& \multicolumn{2}{c}{\textbf{Test}} & \multicolumn{2}{c}{\textbf{Train}} \\
\cmidrule(l){2-3} \cmidrule(l){4-5}
\textbf{\# Tokens} & \textbf{Accuracy} & \textbf{Cosine Sim.} & \textbf{Accuracy} & \textbf{Cosine Sim.} \\
\midrule
1 & 0.3620 & 0.3870 & 0.3884 & 0.4057 \\
2 & 0.3318 & 0.3470 & 0.3391 & 0.3640 \\
5 & 0.2699 & 0.2929 & 0.2753 & 0.2972 \\
10 & 0.1991 & 0.2111 & 0.2055 & 0.2201 \\
\bottomrule
\label{table:forcasting}
\end{tabular}
\end{table*}
\vspace{-100pt}

\begin{table*}[h]
\section{Notations}
\vspace{10pt}
\label{sec:appx.e.}
    \centering
    \renewcommand{\arraystretch}{1.2}
    \resizebox{1.7\columnwidth}{!}{ 
    \begin{tabular}{p{3cm} p{10cm}}
    \hline
    \multicolumn{2}{l}{\textbf{Special Tokens}} \\
    \hline
    $[CLS]$ & Start of sequence token \\
    $[VS], [VE]$ & Start of visit and end of visit tokens \\
    $[REG]$ & Register token \\
    $[W_0]$ to $[W_3]$ & Week tokens (indicating different weeks) \\
    $[M_0]$ to $[M_{12}]$ & Month tokens (indicating different months) \\
    $[LT]$ & Long-term time interval token \\
    $[PAD]$ & Padding token \\
    $[UNK]$ & Unknown token \\
    $[MASK]$ & Mask token \\
    \hline
    \multicolumn{2}{l}{\textbf{Event Tokens and Attributes}} \\
    \hline
    $e_{v,j}$ & Event token at visit $v$, index $j$ \\
    $\tau_{v,j}$ & Timestamp of the event $e_{v,j}$ \\
    $T$ & Type of event (procedure (P), medication (M), or lab result (L)) \\
    $A$ & Age of the patient at the time of the event \\
    $S$ & Visit segment, indicating the separation between adjacent visits \\
    $V$ & Visit order, visit number incrementing for each subsequent visit \\
    $\mathcal{P}$ & Absolute position of the event within the sequence \\
    \hline
    \multicolumn{2}{l}{\textbf{Functions and Mappings}} \\
    \hline
    $f(e_{v,j})$ & Function mapping the event space $\mathcal{E}$ to a set of attributes $\Gamma$ \\
    $\mathcal{C}(v,j)$ & Context of the token $e_{v,j}$ \\
    $\mathcal{M}$ & Set of masked indices for Masked Language Modeling (MLM) \\
    $\mathcal{I}$ & Indices for Next Token Prediction (NTP) task \\
    \hline
    \multicolumn{2}{l}{\textbf{Model Parameters and Objectives}} \\
    \hline
    $\theta, \mathcal{D}$ & Model parameters and Dataset \\
    $y, \hat{y}(x; \theta)$ & Actual and Predicted clinical outcomes \\
    $\mathcal{L}_i$ & Binary cross-entropy loss for task $c_i$ \\
    \hline
    \multicolumn{2}{l}{\textbf{State Space Models (SSMs)}} \\
    \hline
    $\Delta$ & Step size for discretization \\
    $A, B, C$ & Learnable matrices in the State Space Model (SSM) \\
    $h(t)$ & Hidden state in the SSM \\
    $\overline{A}, \overline{B}$ & Discrete parameters derived from continuous parameters \\
    \hline
    \multicolumn{2}{l}{\textbf{Embedding Layer}} \\
    \hline
    $E_{v,j}$ & Embedding of event token $e_{v,j}$ \\
    $E_{\text{concept}}(e_{v,j})$ & Concept embedding of event token $e_{v,j}$ \\
    $E_{\text{type}}(T_{v,j})$ & Token type embedding of event token $T_{v,j}$ \\
    $E_{\text{age}}(A_{v,j})$ & Age embedding of event token $A_{v,j}$ \\
    $E_{\text{time}}(\tau_{v,j})$ & Time embedding of event token $\tau_{v,j}$ \\
    $E_{\text{segment}}(S_{v,j})$ & Segment embedding of event token $S_{v,j}$ \\
    $E_{\text{visit order}}(V_{v,j})$ & Visit order embedding of event token $V_{v,j}$ \\
    $E_{\text{position}}(\mathcal{P}_{v,j})$ & Positional embedding of event token $\mathcal{P}_{v,j}$ \\
    \hline
    \end{tabular}
    }
\end{table*}
\vspace{0pt}
\end{document}